\documentclass{IEEEtran}
\usepackage{amssymb,amsmath,amsthm}
\usepackage{epsfig}
\usepackage{colortbl}
\usepackage[caption=false,font=footnotesize,labelfont=sf,textfont=sf]{subfig}
\usepackage{algorithmic}
\usepackage{algorithm2e}
\usepackage{textcomp}
\usepackage[utf8]{inputenc}
\usepackage[english]{babel}
\usepackage[para]{footmisc}

\begin{document}
\title{Extracting the fundamental diagram from aerial footage}
\author{R.~Makrigiorgis, P.~Kolios, S.~Timotheou, T.~Theocharides, and C.G.~Panayiotou 
\thanks{R.~Makrigiorgis, P.~Kolios, S.~Timotheou, T.~Theocharides, and C.G.~Panayiotou are with the KIOS Research Center for Intelligent Systems and Networks, and the Department of Electrical and Computer Engineering, University of Cyprus, {\tt\small \{makrigiorgis.rafael, pkolios, timotheou.stelios, christosp, ttheocharides\}@ucy.ac.cy}}
}

\maketitle

\begin{abstract}
Efficient traffic monitoring is playing a fundamental role in successfully tackling congestion in transportation networks. Congestion is strongly correlated with two measurable characteristics, the demand and the network density that impact the overall system behavior. At large, this system behaviour is characterized through the fundamental diagram of a road segment, a region or the network.
	
In this paper we devise an innovative way to obtain the fundamental diagram through aerial footage obtained from drone platforms. The derived methodology consists of 3 phases: vehicle detection, vehicle tracking and traffic state estimation. We elaborate on the algorithms developed for each of the 3 phases and demonstrate the applicability of the results in a real-world setting.
\end{abstract}

\section{Introduction}
\label{sec:Introduction}

Drones or Unnamed Aerial Vehicles (UAVs) have a broad range of applications ranging from remote sensing to  deliveries \cite{eurocommision}. They have also become so affordable that they are on a course to transform domains where infrastructure inspection and monitoring in crucial, including of course road traffic monitoring. 

The great advantage of UAVs in road traffic monitoring is that they can capture footage over large areas from which novel information can be extracted. Unlike localize information from loop detectors and static cameras, processed UAV footage can reveal mobility and speed patterns over distances and time periods long enough that the underlying speed-flow-density relationship of lines, road segments and regions can be revealed. More specifically, this speed-flow-density information can be used to extract the fundamental diagram (i.e., the relationship between the traffic flux and the traffic density (vehicles per hour to vehicles per kilometres) \cite{fundamentalD}. It is well know in the transportation research community that this diagram reflects on the macroscopic effects of traffic flux, velocity and density and it often used for predicting the characteristics of the road system behaviour \cite{macroFund}.  Moreover, using the fundamental diagram (FD), traffic control can be applied, such as increasing the road infrastructure at highly congested regions or more favourably apply intelligent traffic light policies and novel traffic managements schemes  as suggested in \cite{papageorgioureview} and looked at in \cite{geroliminis},  \cite{menelaoud}. 

In this paper we elaborate on how the FD can be extracted from video footage collated by UAV platforms through a pipeline of image processing, vehicle tracking and finally, traffic state estimation. Thereafter an example case study will be presented where the pipeline has been implemented and validated using collected GPS traces as well as OBD (Onboard Diagnostic unit) measurements.

The rest of the paper is structured as follows. Section \ref{sec:Related Work} includes related work and demonstrates our contributions with respect to the state-of-the-art. Section \ref{sec:Pipeline} provides a detailed derivation of our proposed pipeline and Section \ref{sec:EvaluationResults} provides an experimental evaluation of this pipeline. Finally Section \ref{sec:Conclusions} concludes with key findings and future research avenues.
	
\section{Related Work}
\label{sec:Related Work}
A plethora of recent works have looked in detail in the problem of road traffic state estimation (using the fundamental diagram) since it largely affects traffic management performance. This is due to the fact that the FD provides a low-complexity modelling framework for characterizing the relationship between  the three main mobility parameters (i.e., speed, flow, and density). Briefly speaking, the FD consists of two distinct regimes that are separated by the critical density of the road traffic infrastructure under investigation. The two regimes are the \textit{free-flow regime} where traffic flows at its maximum speed (i.e., at free-flow speed) and the \textit{congested regime} where traffic experiences speed reduction as density keeps increasing. The concept of the FD has been empirically validated using real traffic data \cite{geroliminisDaganzo} and used to accurately estimate the outflow rate across a road network \cite{daganzo1}. 
 
Traffic control techniques including Gating and Perimeter control, base their policies on the FD to maximize the outflow of a region by controlling its external inflow rate so as the network remains in the free-flow regime \cite{homogeneous, welldefine} \cite{papageorgiou1}, \cite{geroliminis4}. At the same time, Route Guidance methods  aim at balancing the traffic load across the network by selecting routes based on the FD characteristics \cite{geroliminis6}. 

It is therefore evident that an accurate FD model is an essential building block for traffic management. The seminal paper in \cite{macroFund} discusses how GPS traces from a fleet of taxis were used to extract the FD model while the more recent work in \cite{montoya} discusses how the FD can be extracted from scarce sensor data. 

Hereafter, we derive a new and novel approach to extract the FD from aerial video footage that has become both easy and cheap to acquire. Relevant datasets available to date include the Stanford Drone Dataset \cite{stanfordDrone}, which is a dataset having trajectories of multiple road users taken from drone video data. Also, the NGSIM Dataset (Next Generation SIMulation) \cite{ngsim} is a large vehicle dataset, with high-quality traffic data which is destiny to be used in research of traffic flows. There has been evaluations and further analysis of NGSIM dataset in \cite{ngsim, ngsimanalysis} that show however a lot of false positive trajectory collisions and illogical vehicle speeds and accelerations. Specifically for traffic monitoring, the HighD dataset \cite{highd} has recently become available that includes naturalistic vehicle trajectories recorded on German highways. This is a scenario-based testing for the safety validation of highly automated vehicles. HighD also extracts vehicle's trajectory, size and manoeuvres using machine learning and computer vision algorithms. 

Machine learning for detection and tracking of vehicles has been extensively researched in the recent past and our work in \cite{trafficMonKyrkou} is part of that research domain where Convolutional neural networks (CNN) for aerial image processing has been looked at. 

\section{FD Extraction Pipeline}
\label{sec:Pipeline}

As emphasized above, the aim of this work is to provide an end-to-end pipeline for extracting the fundamental diagram form aerial video footage. The three main components of this pipeline are the image processing, vehicle tracking and FD extraction as elaborated below.

\subsection{Image Processing}
Top-down aerial video footage is used as input to a training dataset for vehicle dection. A first step in this procedure is to extract images from the collected videos and either manually annotate vehicles or use tools such as  DronetV3 \cite{dronet}  to automatically annotate images using  templates of the object. In our case a total of $28377$ vehicles were annotated out of $92$ minutes of highway traffic data (captured at $150$m height and covering a road segment of about $130$m length).
	
For evaluating vehicle detections, Darknet YoloV2 \cite{darknetyolov3} and DronetV3 (based on Tiny-Yolov3), were used. As a note, YoloV2 runs in an offline mode while DronetV3 is light-enough to run in real-time.
	
\begin{figure}[b]
	\centering
	\includegraphics[scale=0.8]{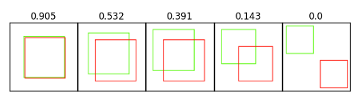}
	\caption{IoU Sample Scores}
	\label{fig:IOUexamples}
\end{figure}

\subsection{Vehicle  Tracking}
Using vehicle detection algorithms each object is pointed out using a bounding box with IDs that change over time due to the lack of accurate data association. To address this problem and be able to track correctly vehicle trajectories, in this work the Hungarian Algorithm \cite{hungarianAlgo} in combination with Kalman filtering \cite{kalmanFilter} is used. 

By employing the Hungarian algorithm (also known as Kuhn-Munkres Algorithm), an object in the current frame is matched to an object in the previous frame using a score function. To associate objects in consecutive frames, the IoU (Intersection of Union) is employed here where the percentage of overlap between frames is used as exemplified in Fig. \ref{fig:IOUexamples}. When the IoU scores above a certain threshold are found, matching the previous bounding box with the current detected box results to a good representation of bounding box trajectories for each vehicle. 

Evidently, the performance of this approach degrades when  vehicle dynamically change speeds or take sharp turns or even when an occlusion occurs (eg. a vehicle passing under a tree). To address the aforementioned cases, on top of the matching between successive frames, Kalman filtering is also employed.

\begin{figure}[h]
	\centering
	\includegraphics[width=2.5in, trim=5mm 5mm 15mm 3mm]{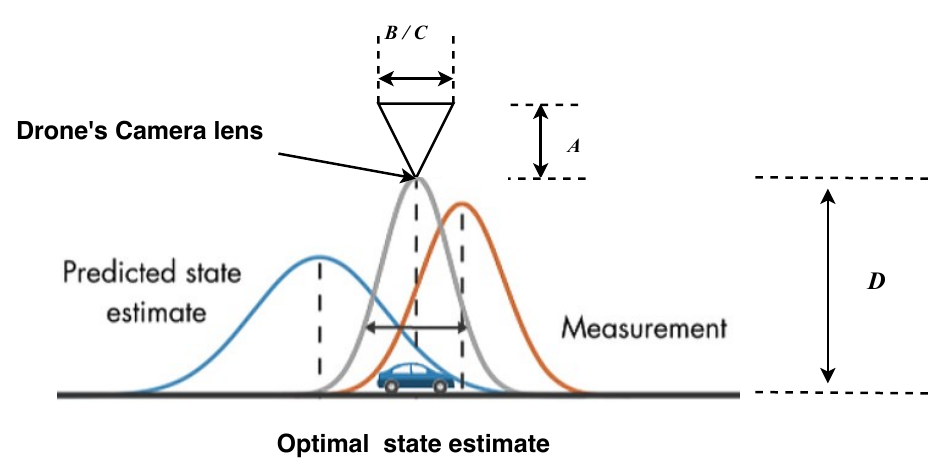}
	\caption{Kalman Filter Explanation}
	\label{fig:kalman_filter}
	\end{figure}

Kalman filtering is applied on every bounding box after a box has been matched using the Hungarian algorithm. When the association is made, predictions and corrections (updating Kalman equations with real measurements) are made. To calculate the mean and covariance values, OpenCV's Kalman Filter library is used. An example of what does Kalman Filter actually calculate is shown in Fig. \ref{fig:kalman_filter}. 

\begin{figure}[h]
	\centering
	\includegraphics[scale=0.195, trim=5mm 10mm 15mm 10mm]{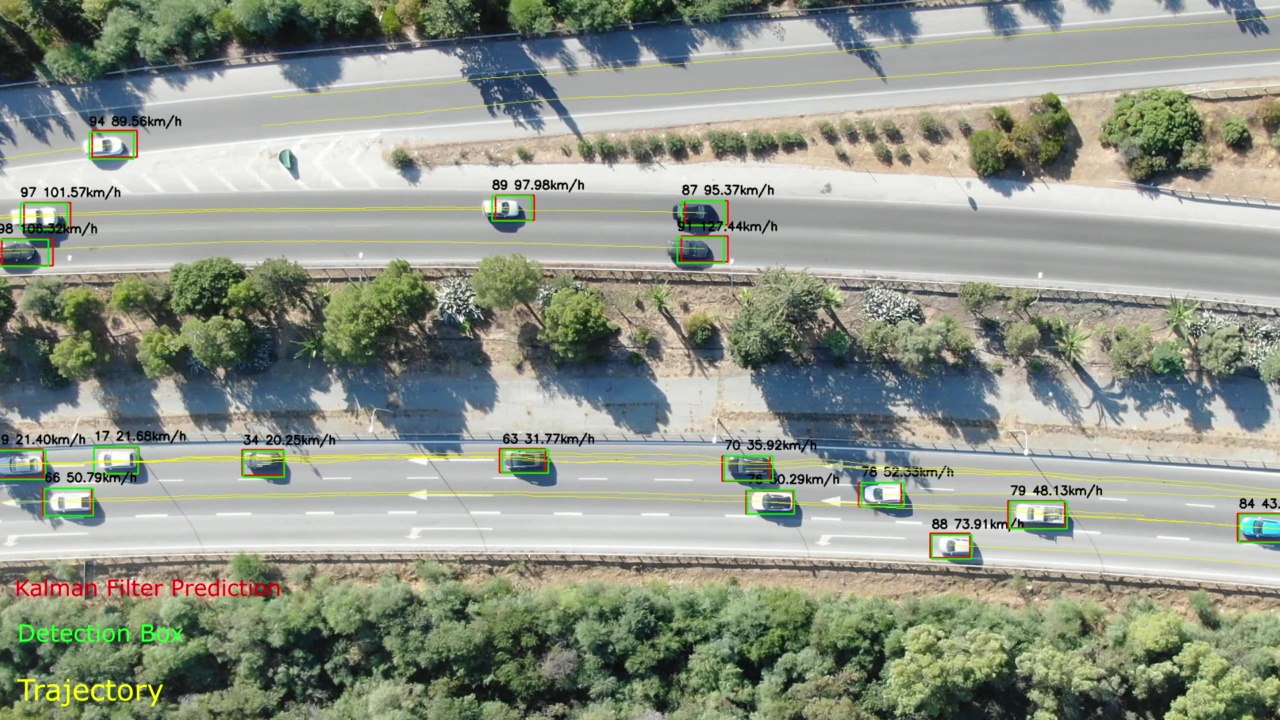}
	\caption{Harpy's Kalman Filter usage preview (best viewed in colour)}
	\label{fig:kalman_filter_prev}
	\end{figure}

In essence Kalman filtering is employed to keep track of every vehicle crossing in the field-of-view. In those cases where a vehicle dynamically changes speeds or positions the IoU of the boxes between two frames may differ in such a way that it cannot be matched as the same vehicle. Instead, using Kalman filtering, predictions of the detected boxes (as shown in Fig. \ref{fig:kalman_filter_prev}) and vehicle tracking becomes much more accurate. A pseudocode of the proposed approach can be found in Alg. \ref{alg:harpy_track}.

\setlength{\textfloatsep}{3pt}

\subsection{Addressing Occlusion}
When an occlusion occurs, vehicle stop being observed by the camera. In this case, the Kalman filter can still predict the next position of the vehicle, use that information to track the vehicle trajectory and eventually match up with the detected vehicle when it becomes visible again. 

The main challenge here is the fact that since Kalman is not updating its measurements, a motion model needs to be introduced. Hereafter we use a simple linear motion model where the displacement between the last set of frames where the vehicle was detected is used to estimate subsequent vehicle positions.

\begin{figure}[h]
	\centering
	\includegraphics[scale=0.42, trim=5mm 5mm 15mm 3mm]{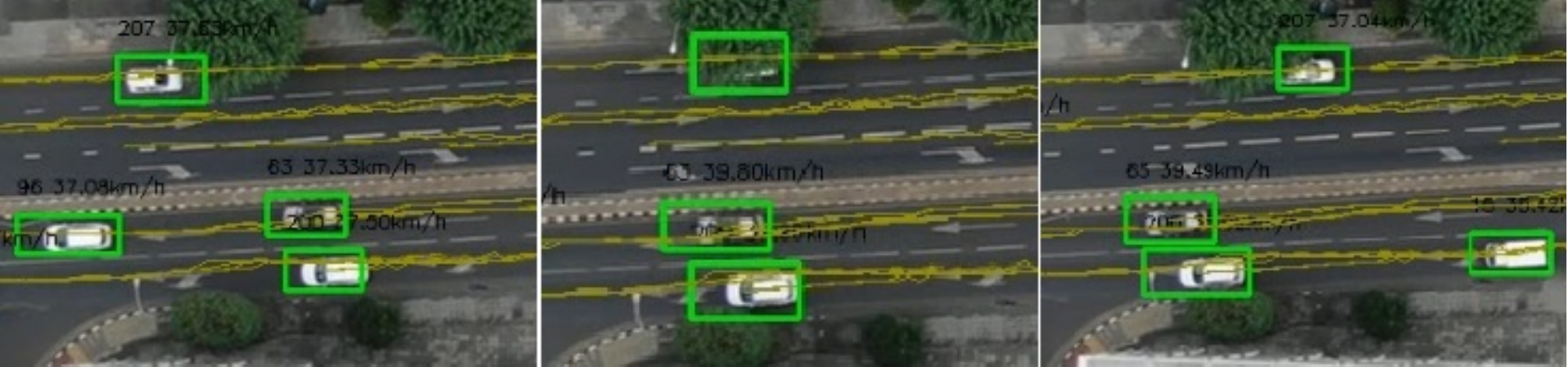}
	\caption{Harpy's Kalman Filter usage for occlusion from a tree (ID 207 of the vehicle remains the same).}
	\label{fig:kalman_filter_occlusion}
\end{figure}

To aid understanding, Fig. \ref{fig:kalman_filter_occlusion} provides an illustrative example. In case the predicted vehicle trajectory does not match with a detected vehicle over a certain period of time, the estimates are discarded. Clearly, predictions made over extended periods of time will substantially deviate from reality due to the model imperfections. A pseudocode of the occlusion algorithm can be found in Algorithm \ref{alg:Occlusion_pseudo}.
	\begin{algorithm}
	\label{alg:harpy_track}
	\SetAlgoLined
	\KwResult{Vehicle Bounding Box and Statistics }
	\While{Boxes detected previously}{
		calculate IoU with current detection\;
		\eIf{best IoU score exists and bigger than threshshold}{
			match boxes - previous with current box\;
		}{
			previous box not found\;
		}
	}
	\eIf{previous box not found}{
		initialization of box\;
	}{
		Calculate average of Euclidean distance from previous frames to current (only for the latest $25$ frames)\;
		Calculate velocity of the vehicle in Km/h \;
		Kalman Predict()\;
		Kalman Correct() using real mesurements\;
		Calculate Direction of the vehicle - depending on the box\;
		Display Bounding Box and Trajectory\;
		Save all the statistics of the vehicle to array\;
	}

	\caption{Harpy Detection and Tracking Algorithm}
\end{algorithm}

	\setlength{\textfloatsep}{3pt}
	\begin{algorithm}[t]
	\label{alg:Occlusion_pseudo}
	\SetAlgoLined
	\KwResult{Display Vehicle Bounding Box on Occlusion }
	\eIf{Vehicle stop being detected for less than X frames}{
		\eIf{is new detection}{
			ignore it\;
		}{
			Kalman Predict()\;
			Calculate x,y difference from previous frames\;
			Kalman Correct()\;
			Display the Box\;	
		}	
	}{
		remove the box from being active\;	
	}
	\caption{Harpy Occlusion prevention }
	\end{algorithm}

\subsection{Velocity Estimation}
To calculate the velocity of moving vehicles from detections, a representation of pixels to real distances is needed. To obtain that relationship, the Ground Sample Distance (GSD) is employed as mentioned in \cite{propello}. Ground Sample Distance is the distance between centre points of each sample taken of the ground. In simpler terms, the GSD is the representation, in real size, of each pixel on the 2D plane. Calculating it requires a set of parameters such as the UAV height, the camera's sensor height and width, focal length of the camera and the image width, height of the video taken. Of course, these parameters need to be adjusted when either image size, UAV height or camera lenses are changed. The latter parameters can be taken from the manufacturers technical specifications. Then, by calculating the GSD for height and width separately, the worst case scenario is picked as our GSD. The equations are as follows:
	\[
	GSD_{h}=\frac{D * B}{A * E} , ~~~~GSD_{w}=\frac{D * C}{A * F}
	\]

	\[
	GSD_{final}=\frac{GSD_{worst}}{1000}=\frac{Km}{ pixels}
	\]
	Assuming A is the focal length, B and C are the camera's sensor height and width respectively, D is the drone's height and E, F are the image's height and width respectively.  A showcase of what these parameters are is shown in Fig. \ref{fig:kalman_filter}.

When GSD is calculated a correct representation of centimetres to pixels (cm/px) is obtained and used to calculate the average Euclidean distance over consecutive frames. To calculate the average of the Euclidean, the difference between the last $f$ detected frames is accounted for. Then given the frame rate of the video the velocity for each vehicle trajectory can be calculated as follows: 
	\[
	{\textstyle\hspace*{0.1cm}Eu=\sqrt{(|x2-x1|*GSD)^{2}+(|y2-y1|*GSD)^{2}}}
	\]

	\[
		Velocity=\frac{(\sum_{1}^{25}Eu)*FR}{FD}\>\left(\frac{Km}{h}\right)
	\]
assuming FR and FD are Frame Rate and Frame Difference respectively and Eu is the Euclidean. 

\begin{figure}[h]
		\centering
		\includegraphics[width=3in]{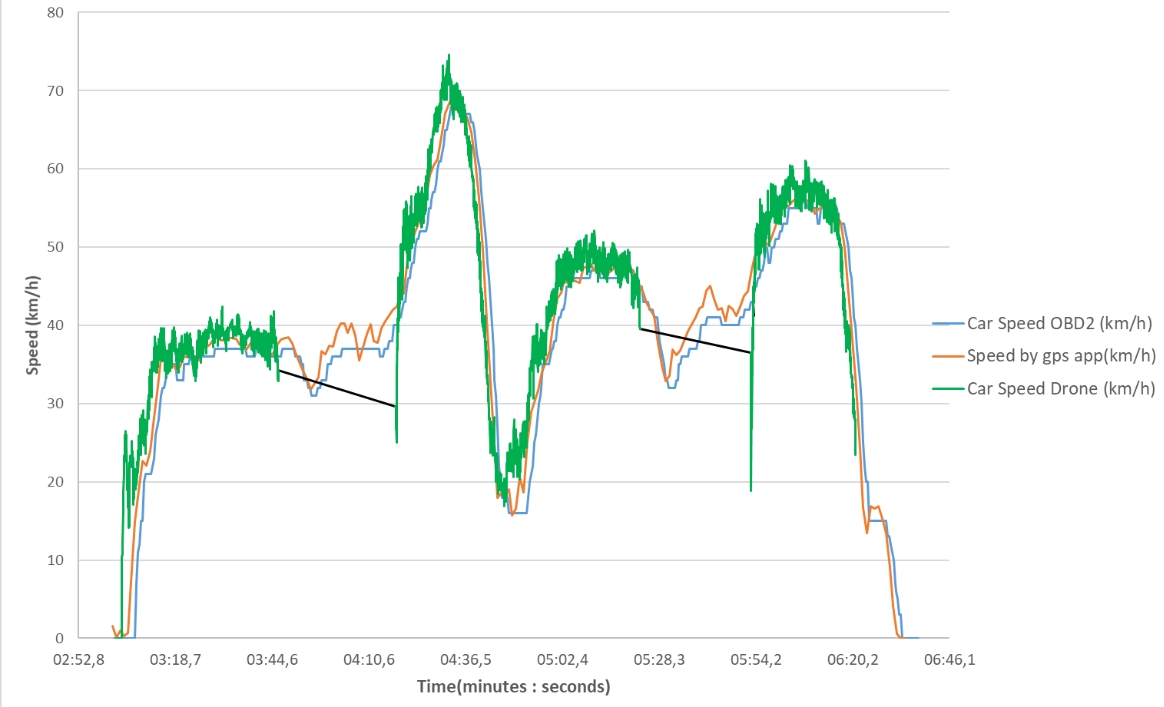}		
		\caption{Comparison of actual speed measurements captured by the three complimentary approaches.  A small offset in the timing of the measurements is related to matching the video frames to the timestamped data collected from OBD and GPS trackers.}
		\label{fig:velocityConfirm}
\end{figure}

To verify these estimates, a simple real-life experiment was conducted using a test vehicle. Aerial footage of our test vehicle was collected while driving over a particular road segment. Video recording were made using different heights between $50$ to $500$ meters. At the same time an OBD (onboard diagnostic unit) was used to capture timestamped readings of the vehicle speed while a GPS tracker was used in order to take measurements of the position and hence the velocity of the vehicle as well. As it turns out from the comparison of these three methods, the proposed tracking algorithm was able to achieve an accuracy of $90\%$ in the speed estimates as compared to the OBD. The experiment test results for the case of the 150m height data acquisition can be seen in Fig. \ref{fig:velocityConfirm}.	The straight lines connecting the traces acquired from the aerial footage account for the time periods where the vehicle went out and back in the field-of-view of the UAV during the experiment.

\subsection{Traffic Monitoring Statistics}
In addition to velocity estimates, vehicle detections can also provide a number of additional measurements including per frame vehicle density and inflow/outflow vehicle counters. In effect, this information can be used to extract the fundamental diagram of a road segment and be used to characterize the traffic state. In summary, the following set of data were extracted using the proposed pipeline:
\begin{itemize}
	\item X,Y position and timestamp of every vehicle for every detection.
	\item Vehicle Velocities for each detection.
	\item Vehicle directions for each detection (left, right, top/bottom-right/left) based on the boxes difference between each frame.
	\item Density of vehicles in each frame.
	\item Inflow/outflow of vehicles.
\end{itemize}
   
\section{Experimental Evaluation}
\label{sec:EvaluationResults}

To demonstrate the applicability of the Harpy FD pipeline, 3 hours of aerial video was captured using DJI Mavic Enterprise UAVs flying at 150m altitude above a single road segment in Nicosia, Cyprus. The training was done using more than $1000$ images with more than $28000$ vehicle annotations obtained from various own and online sources. Furthermore, both YoloV2 and DronetV3 networks were employed for performance comparison. The main reason of using DronetV3 is to investigate the trade-off between processing time and detection accuracy. As a note, even though the data was being processed offline, having a smaller network can significantly reduce the processing time, especially when dealing with long duration and high quality video footage. Also another reason for choosing YoloV2 instead of YoloV3 for offline detection was due to the fact that the system configuration could for example, not be able to handle detection of $2K$ (or higher) resolution footage using YoloV3 due to lack of memory resources. Using YoloV2, we could extract detections of $2K$ resolution (or downscaling a $4k$ video to $2K$) footage and since results were obtained from that process it was also considered. 

Taking a look on Table \ref{table:iouMap} it is clear that in terms of MAP (Mean Average Precision) accuracy of these two networks, they are not too much apart, since YoloV2 does not have much more layers than DronetV3. Although, the IoU percentage of the ground truth of the detections is much more higher using YoloV2. That is another reason why YoloV2 was chosen for our Harpy Dataset example. Having a better IoU means that the boxes of the detections are much more accurate in terms of vehicle shapes and resulting to more precise trajectories.

The training and detection tests were done using a desktop computer with an i7-7800X 12 core CPU @3.5Ghz, 64GB of RAM and an NVIDIA RTX 2080 11GB. The evaluation was done on the collected 3 hour video where more than $15000$ vehicles were extracted. 
	
\begin{table}[h]		
	\caption{Table of IoU and MAP comparison of Dronet and Yolo.}
	\label{table:iouMap}	\scalebox{0.78}{%

	\begin{tabular}{|c| c| c| c |c | c| c| c| c|}
		\hline			 
		&\multicolumn{4}{c|}{IOU (\%) } &\multicolumn{4}{c|}{MAP (\%)} \\
		\hline
		Threshold(\%)& $15$ &	$25$ & $50$ & $75$ & $15$ &	$25$ & $50$ & $75$\\ 
		\hline
		\textbf{YoloV2} &	$75,46$&	$75,38$&	$73,34$ &	$63,51$	&	$44,89$ &	$44,88$	&	$40,54$	&	$33,51$\\
		\hline
		DronetV3 &	$44,82$&	$51,89$&	$48,69$ &	$21,54$	&	$38,53$ &	$41,68$	&	$35,89$	&	$10,68$\\
		\hline
	\end{tabular}
	} 
\end{table}

\begin{figure}[h]
		\centering
		\includegraphics[width=3in]{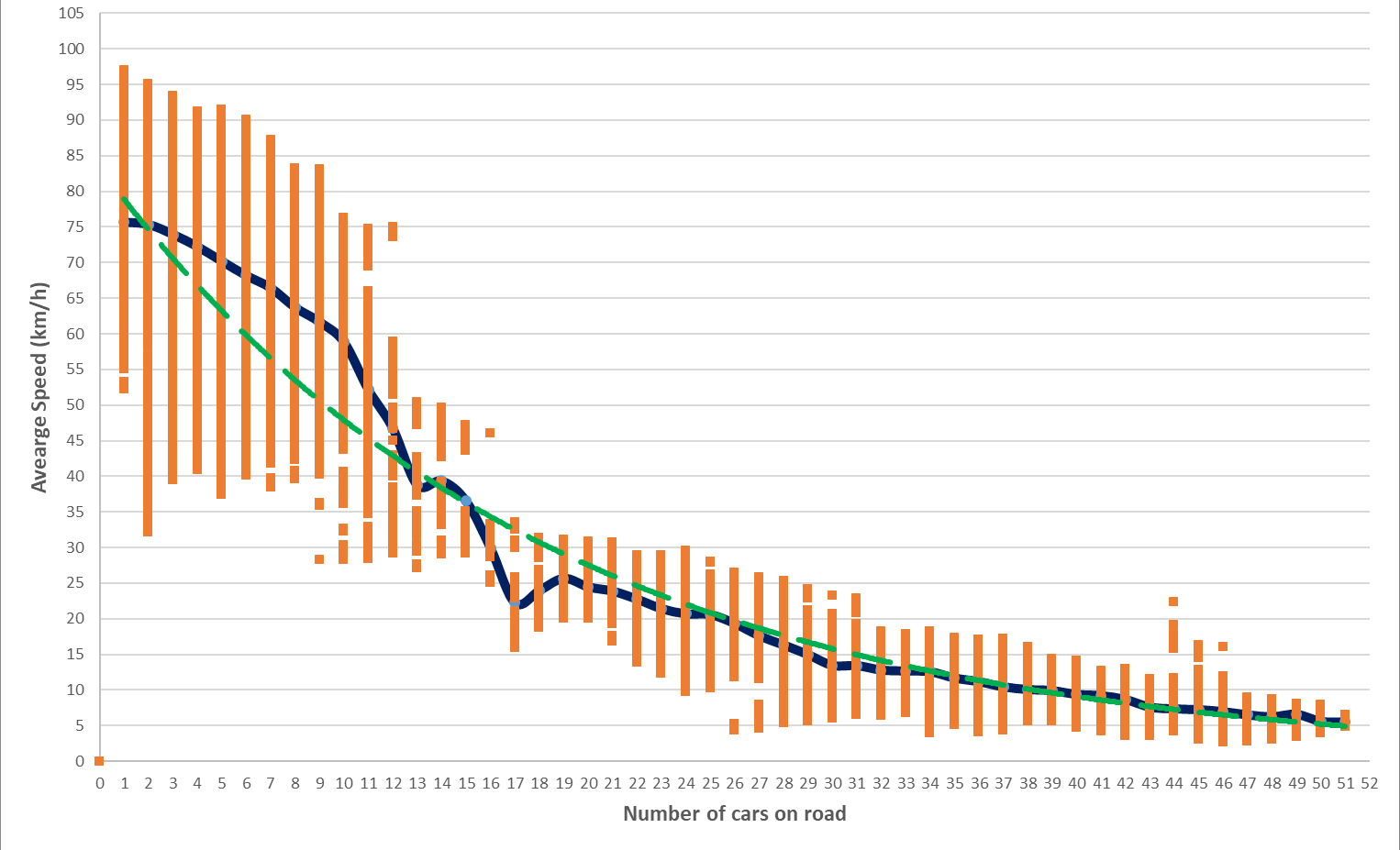}
		\caption{Speed-density relationship extracted using the Harpy dataset. }
		\label{fig:avgspeedGraph}
\end{figure}

With respect to the obtained vehicle trajectories, Fig. \ref{fig:avgspeedGraph} plots the speed-density diagram showing with orange the actual values per frame, with blue the average speed per density value and with green the approximate non-linearly relationship of speed with density.

\begin{figure}[h]
		\centering
		\includegraphics[width=3in]{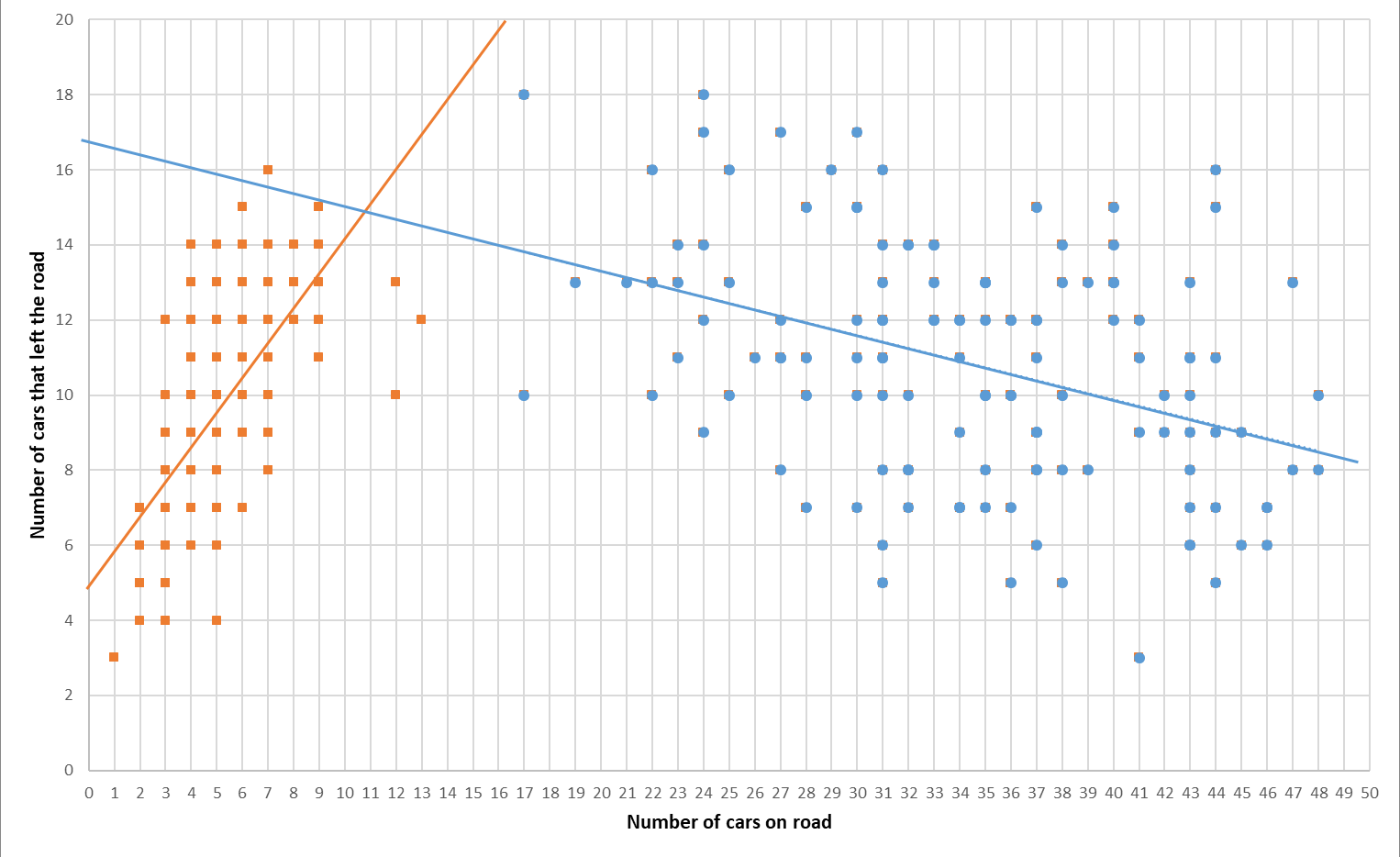}
		\caption{Fundamental Diagram. It represents the relationship between traffic flux to traffic density. When the FD graph decays, it means traffic jam starts to occur. }
		\label{fig:FDGraph}
\end{figure}

The FD is depicted in Fig. \ref{fig:FDGraph}. As shown in the plot, the critical density region is between 10-20 vehicles below which the road segments experiences free-flow conditions and above that congestion appears. In the latter case, the traffic flux decreases causing traffic congestion. 
	
The Harpy Dataset example videos, CSV files and exported diagrams can be found in our website at https://www.kios.ucy.ac.cy/harpydata. 
	
\section{Conclusion and Future Work}
\label{sec:Conclusions}
This work develops a detailed pipeline for traffic monitoring using aerial video data through detection and tracking of vehicles and finally traffic state estimation. The proposed pipeline was empirically evaluated using OBD and GPS measurements. Thereafter the proposed pipeline was used to extract the FD from video collected from a particular road section in Nicosia, Cyprus.

As future work, we will be exploring online solutions based on light-weight deep learning algorithms (e.g., \cite{efficientDeepLearning2017}) that could provide adequate accuracy and run in real-time on resource-limited onboard UAV processors. In addition, our aim is to explore our solution over complementary datasets with varying parameters. 

\section*{Acknowledgement}
This work is supported by the European Union’s Horizon 2020 research and innovation programme under grant agreement No 739551 (KIOS CoE) and from the Republic of Cyprus through the Directorate General for European Programmes, Coordination and Development.

\bibliographystyle{ieeetran}

\bibliography{Harpy_paper}

\begin{thebibliography}{10}
\providecommand{\url}[1]{#1}
\csname url@samestyle\endcsname
\providecommand{\newblock}{\relax}
\providecommand{\bibinfo}[2]{#2}
\providecommand{\BIBentrySTDinterwordspacing}{\spaceskip=0pt\relax}
\providecommand{\BIBentryALTinterwordstretchfactor}{4}
\providecommand{\BIBentryALTinterwordspacing}{\spaceskip=\fontdimen2\font plus
\BIBentryALTinterwordstretchfactor\fontdimen3\font minus
  \fontdimen4\font\relax}
\providecommand{\BIBforeignlanguage}[2]{{%
\expandafter\ifx\csname l@#1\endcsname\relax
\typeout{** WARNING: IEEEtran.bst: No hyphenation pattern has been}%
\typeout{** loaded for the language `#1'. Using the pattern for}%
\typeout{** the default language instead.}%
\else
\language=\csname l@#1\endcsname
\fi
#2}}
\providecommand{\BIBdecl}{\relax}
\BIBdecl

\bibitem{eurocommision}
E.~Commission, ``Commission staff working document - towards a european
  strategy for the development of civil applications of remotely piloted
  aircraft systems(rpas),'' 2012.

\bibitem{fundamentalD}
G.~Puppo, M.~Semplice, A.~Tosin, and G.~Visconti, ``Fundamental diagrams in
  traffic flow: the case of heterogeneous kinetic models,'' \emph{arXiv
  preprint arXiv:1411.4988}, 2014.

\bibitem{macroFund}
N.~Geroliminis and C.~F. Daganzo, ``Existence of urban-scale macroscopic
  fundamental diagrams: Some experimental findings,'' \emph{Transportation
  Research Part B: Methodological}, vol.~42, no.~9, pp. 759--770, 2008.

\bibitem{papageorgioureview}
M.~Papageorgiou, C.~Diakaki, V.~Dinopoulou, A.~Kotsialos, and Y.~Wang, ``Review
  of road traffic control strategies,'' \emph{Proceedings of the IEEE},
  vol.~91, no.~12, pp. 2043--2067, 2003.

\bibitem{geroliminis}
A.~Kouvelas, M.~Saeedmanesh, and N.~Geroliminis, ``Enhancing model-based
  feedback perimeter control with data-driven online adaptive optimization,''
  \emph{Transportation Research Part B: Methodological}, vol.~96, pp. 26--45,
  2017.

\bibitem{menelaoud}
C.~Menelaou, S.~Timotheou, P.~Kolios, and C.~Panayiotou, ``Joint route guidance
  and demand management for multi-region traffic networks,'' pp. 2183--2188,
  2019.

\bibitem{geroliminisDaganzo}
N.~Geroliminis, C.~F. Daganzo \emph{et~al.}, ``Macroscopic modeling of traffic
  in cities,'' no. 07-0413, 2007.

\bibitem{daganzo1}
C.~Daganzo, ``Urban gridlock: Macroscopic modeling and mitigation approaches,''
  \emph{Transportation Research Part B: Methodological}, vol.~41, pp. 49--62,
  01 2007.

\bibitem{homogeneous}
A.~Mazloumian, N.~Geroliminis, and D.~Helbing, ``The spatial variability of
  vehicle densities as determinant of urban network capacity,''
  \emph{Philosophical Transactions of the Royal Society A: Mathematical,
  Physical and Engineering Sciences}, vol. 368, no. 1928, pp. 4627--4647, 2010.

\bibitem{welldefine}
N.~Geroliminis and J.~Sun, ``Properties of a well-defined macroscopic
  fundamental diagram for urban traffic,'' \emph{Transportation Research Part
  B: Methodological}, vol.~45, no.~3, pp. 605--617, 2011.

\bibitem{papageorgiou1}
M.~Keyvan-Ekbatani, A.~Kouvelas, I.~Papamichail, and M.~Papageorgiou,
  ``Exploiting the fundamental diagram of urban networks for feedback-based
  gating,'' \emph{Transportation Research Part B: Methodological}, vol.~46,
  no.~10, pp. 1393--1403, 2012.

\bibitem{geroliminis4}
J.~Haddad and N.~Geroliminis, ``On the stability of traffic perimeter control
  in two-region urban cities,'' \emph{Transportation Research Part B:
  Methodological}, vol.~46, no.~9, pp. 1159--1176, 2012.

\bibitem{geroliminis6}
M.~Yildirimoglu and N.~Geroliminis, ``Approximating dynamic equilibrium
  conditions with macroscopic fundamental diagrams,'' \emph{Transportation
  Research Part B: Methodological}, vol.~70, pp. 186--200, 2014.

\bibitem{montoya}
O.~Q. Montoya and C.~Canudas-de Wit, ``Estimation of fundamental diagrams in
  large-scale traffic networks with scarce sensor measurements,'' pp.
  3457--3462, 2018.

\bibitem{stanfordDrone}
A.~Robicquet, A.~Sadeghian, A.~Alahi, and S.~Savarese, ``Learning social
  etiquette: Human trajectory understanding in crowded scenes,'' pp. 549--565,
  2016.

\bibitem{ngsim}
B.~Coifman and L.~Li, ``A critical evaluation of the next generation simulation
  (ngsim) vehicle trajectory dataset,'' \emph{Transportation Research Part B:
  Methodological}, vol. 105, pp. 362--377, 2017.

\bibitem{ngsimanalysis}
M.~Montanino and V.~Punzo, ``Trajectory data reconstruction and
  simulation-based validation against macroscopic traffic patterns,''
  \emph{Transportation Research Part B: Methodological}, vol.~80, pp. 82--106,
  2015.

\bibitem{highd}
R.~Krajewski, J.~Bock, L.~Kloeker, and L.~Eckstein, ``The highd dataset: A
  drone dataset of naturalistic vehicle trajectories on german highways for
  validation of highly automated driving systems,'' pp. 2118--2125, 2018.

\bibitem{trafficMonKyrkou}
C.~Kyrkou, S.~Timotheou, P.~Kolios, T.~Theocharides, and C.~G. Panayiotou,
  ``Optimized vision-directed deployment of uavs for rapid traffic
  monitoring,'' pp. 1--6, 2018.

\bibitem{dronet}
C.~Kyrkou, G.~Plastiras, T.~Theocharides, S.~I. Venieris, and C.-S. Bouganis,
  ``Dronet: Efficient convolutional neural network detector for real-time uav
  applications,'' pp. 967--972, 2018.

\bibitem{darknetyolov3}
J.~Redmon and A.~Farhadi, ``Yolov3: An incremental improvement,'' \emph{arXiv
  preprint arXiv:1804.02767}, 2018.

\bibitem{hungarianAlgo}
H.~W. Kuhn, ``The hungarian method for the assignment problem,'' \emph{Naval
  research logistics quarterly}, vol.~2, no. 1-2, pp. 83--97, 1955.

\bibitem{kalmanFilter}
R.~E. Kalman, ``A new approach to linear filtering and prediction problems,''
  \emph{Journal of basic Engineering}, vol.~82, no.~1, pp. 35--45, 1960.

\bibitem{propello}
P.~Aero, ``What is ground sample distance (gsd) and how does it affect your
  drone data?'' \emph{Propeller February}, 2018.

\bibitem{efficientDeepLearning2017}
V.~Sze, Y.-H. Chen, T.-J. Yang, and J.~S. Emer, ``Efficient processing of deep
  neural networks: A tutorial and survey,'' \emph{Proceedings of the IEEE},
  vol. 105, no.~12, pp. 2295--2329, 2017.

\end{thebibliography}

\end{document}